%% file: main.tex
\documentclass{llncs}

\usepackage[utf8]{inputenc}
\usepackage{amsmath}
\usepackage{graphicx}
\usepackage{comment}
\usepackage{amsfonts}
\usepackage{amssymb}
\usepackage{subcaption}
\usepackage{url}
\usepackage{listings}
\lstdefinelanguage{json}{
    basicstyle=\normalfont\ttfamily,
    commentstyle=\color{black}, 
    stringstyle=\color{black}, 
    numbers=left,
    numberstyle=\scriptsize,
    stepnumber=1,
    basicstyle=\tiny,
    numbersep=8pt,
    showstringspaces=false,
    breaklines=true,
    frame=lines,
    numbers=none
    backgroundcolor=\color{white}, 
    string=[s]{"}{"},
    comment=[l]{:\ "},
    morecomment=[l]{:"},
    literate=
        *{0}{{{\color{black}0}}}{1}
         {1}{{{\color{black}1}}}{1}
         {2}{{{\color{black}2}}}{1}
         {3}{{{\color{black}3}}}{1}
         {4}{{{\color{black}4}}}{1}
         {5}{{{\color{black}5}}}{1}
         {6}{{{\color{black}6}}}{1}
         {7}{{{\color{black}7}}}{1}
         {8}{{{\color{black}8}}}{1}
         {9}{{{\color{black}9}}}{1}
}

\begin{document}

{\let\thefootnote\relax\footnotetext{Copyright \textcopyright\ 2020 for this paper by its authors. Use permitted under Creative Commons License Attribution 4.0 International (CC BY 4.0). CLEF 2020, 22-25 September 2020, Thessaloniki, Greece.}}

\title{Unsupervised Pre-training for Biomedical Question Answering}

\newcommand*\samethanks[1][\value{footnote}]{\footnotemark[#1]}

\author{Vaishnavi Kommaraju\thanks{Equal Contribution} \inst{1} \and Karthick Gunasekaran\samethanks \inst{1} \and
Kun Li\samethanks \inst{1}
\and Trapit Bansal \inst{1} \and Andrew McCallum \inst{1} \and
Ivana Williams \inst{2} \and 
Ana-Maria Istrate \inst{2}} 
\authorrunning{F. Author et al.}
%
\institute{University of Massachusetts Amherst, USA \and
Chan Zuckerberg Initiative, USA\\
\email{\{vkommaraju, kgunasekaran, kunli, tbansal, mccallum\}@cs.umass.edu}\\
\email{\{iwilliams, aistrate\}@chanzuckerberg.com}}

\maketitle

\begin{abstract}
\input{abstract}
\end{abstract}

\section{Introduction}
\input{introduction.tex}

\section{Model}
\input{model.tex}

\section{Related Work}
\input{related_work.tex}

\section{Experiments}
\input{experiments.tex}

\section{Conclusion}
\input{conclusion.tex}

\section{Acknowledgements}
\input{acknowledgement.tex}

 \bibliographystyle{splncs04}
 \bibliography{references}

\end{document}

%% file: abstract.tex
We explore the suitability of unsupervised representation learning methods on biomedical text -- BioBERT, SciBERT, and Bio-SentVec -- for biomedical question answering.
To further improve unsupervised representations for biomedical QA, we introduce a new pre-training task from unlabeled data designed to reason about biomedical entities in the context.
Our pre-training method consists of corrupting a given context by randomly replacing some mention of a biomedical entity with a random entity mention and then querying the model with the correct entity mention in order to locate the corrupted part of the context.
This de-noising task enables the model to learn good representations from abundant, unlabeled biomedical text that helps QA tasks and minimizes the train-test mismatch between the pre-training task and the downstream QA tasks by requiring the model to predict spans.
Our experiments show that pre-training BioBERT on the proposed pre-training task
significantly boosts performance and outperforms the previous best model from the 7th BioASQ Task 7b-Phase B challenge.

\keywords{Biomedical question answering \and self-supervised learning \and language models}

%% file: introduction.tex
Accurate systems for biomedical question-answering
have the potential to be useful for a range of problems including clinical decision making,
researching disease treatments and symptoms, 
answering user questions and more.
The task of machine reading comprehension (MRC) aims to evaluate this ability, where the model is presented with a context along with a question regarding the context and is expected to predict the answer to the question.
MRC has received significant interest, where specially in the general domain several large datasets for supervised learning of MRC models have been proposed \cite{liu2019neural,zeng2020survey}.
Recently, self-supervised pre-training of transformer models \cite{devlin2019bert,radford2019language,yang2019xlnet} with language modeling objectives has been shown to learn good feature representations and improve performance on many question-answering tasks \cite{devlin2019bert,joshi2019spanbert,keskar2019unifying}.

In the biomedical domain, large MRC datasets have been scarce as annotating data can be expensive, and typical datasets, for example training dataset of BioASQ competitions \cite{tsatsaronis2012bioasq}, are significantly smaller than general domain datasets like SQuAD \cite{Rajpurkar_2016}, which limits the accuracy of supervised models for biomedical QA.
To overcome this, several recent methods have leveraged different avenues of distant supervision to create large datasets for learning MRC models \cite{Pampari2018emrQAAL,Jin2019PubMedQAAD}. 
Moreover, following the success of unsupervised pre-training \cite{devlin2019bert},
recent methods have pre-trained BERT language model on biomedical datasets \cite{lee2019biobert,beltagy2019scibert,alsentzer2019publicly}, which has been shown to learn better representations for biomedical text, improving performance for biomedical MRC \cite{lee2019biobert,yoon2019pretrained}.

In this work, we focus on learning good representations of biomedical text from unsupervised data that are helpful for QA.
To this end, we consider three popular methods for unsupervised representations in the biomedical domain: BioBERT \cite{lee2019biobert}, SciBERT \cite{beltagy2019scibert}, and BioSentVec \cite{Chen2018BioSentVecCS}.
We evaluate the performance of these methods with fine-tuning on three MRC tasks: factoid, list and yes/no questions from the BioASQ Task 8b challenge.
Further, since transfer learning from general domain QA datasets has been useful for improving performance on biomedical QA \cite{wiese2017neural,lee2019biobert}, we evaluate transfer from SQuAD \cite{Rajpurkar_2016} and PubMedQA \cite{Jin2019PubMedQAAD} datasets when using the pre-trained models and find improvements when using these datasets for additional fine-tuning.
Finally, to leverage abundant unlabelled biomedical data, 
we develop a new pre-training method for improving biomedical MRC performance.

Our pre-training method is focused on learning good representations of biomedical text and developing reasoning about entities in context to help MRC tasks. 
The pre-training approach involves finding mentions of entities, using a biomedical named entity tagger \cite{wei2013pubtator}, and corrupting a random entity mention in the context by replacing it with another random entity mention from the corpus. The model is then queried with the correct entity mention -- similar to a question in an MRC model -- and is required to predict the location of the corrupted entity mention from the context.
As the method does not require expensive human annotation it benefits from training on large amounts of biomedical text data. We use a large corpora of Pubmed abstracts for this pre-training.
Moreover, the pre-training approach minimizes train-test mismatch for MRC and we can reuse the entire MRC model used in pre-training (including classification layers) and fine-tune it for any particular biomedical MRC task.

We evaluate these approaches on the tasks of factoid, yes/no and list questions using BioASQ 7b challenge dataset and submit the trained models for the BioASQ Task 8b Biomedical Semantic QA challenge.
The main observations from this work include:
\begin{enumerate}
\item Self Supervised de-noising approach improves the performance for all three question types.
\item  Performance of BioBERT and SciBERT is observed to be comparable.
\item Using general domain QA data, such as SQuAD and PubmedQA, for additional fine-tuning of the pre-trained model improves performance on biomedical QA.
\item  BioSentVec can be used to supplement BioBERT/SciBERT model's performance but doesn't perform well on its own.
\end{enumerate}

We describe the modeling approach that we consider as well as the pre-trained models used in Section \ref{sec:model}, describe the proposed self-supervised de-noising approach in Section \ref{sec:unsup}, present our experimental results and analysis in Section \ref{sec:experiments}, discuss related work in Section \ref{sec:related}, and conclude in Section \ref{sec:conclusion}.

%% file: model.tex
\label{sec:model}
We discuss here the pre-trained models considered and the QA model used for the three types of questions.

\subsection{Pre-trained models: BioBERT and SciBERT}
We evaluate the performance of pre-trained language models- BioBERT \cite{lee2019biobert} and SciBERT \cite{beltagy2019scibert} for our task. 
BioBERT and SciBERT are transformer \cite{vaswani2017attention} based models. The input to these is the tokenized question concatenated with corresponding passage (either abstract or snippet) using a separator token \cite{devlin2019bert}. The input is also prefixed with a special CLS token which can be used as a sentence representation.
The representation of each token in input is composed of the concatenation of the embeddings for the token, segment, and its position.
These embeddings are then passed through multiple layers of self-attention which yield contextualized representations for each token of the input.
For factoid and list type questions we utilize these contextualized token representations whereas yes/no questions utilize the CLS representation from the final layer. 
\paragraph{}
\textbf{BioBERT} \cite{lee2019biobert} model was the first BERT model \cite{devlin2019bert} trained on biomedical text using the pre-training method introduced by BERT \cite{devlin2019bert}.
It is pre-trained on 18B words of PubMed (from abstracts and full text articles) apart from Wikipedia and Books corpus originally used in the BERT training.
BioBERT largely outperforms BERT and previous state-of-the-art models in a variety of biomedical text mining tasks when pre-trained on the biomedical corpora. In three representative biomedical NLP tasks including biomedical named entity recognition, relation extraction, and question answering, BioBERT outperforms most of the previous state-of-the-art models. 
\paragraph{}
\textbf{SciBERT} \cite{beltagy2019scibert} was trained on papers from the corpus of semanticscholar.org. The corpus size was 1.14M research papers with 3.1B tokens and uses the full text of the papers in training, not just abstracts. SciBERT has its own vocabulary (scivocab) that's built to best match the training corpus. The training procedure of SciBERT is similar to BioBERT.


\subsection{Question answering model}
Our text representations come from BioBERT or SciBERT models. There are then passed through task-specific layers for each QA task: yes/no, factoid and list. 
The weights of the task-specific layers as well as the BioBERT/SciBERT weights are all fine-tuned during the training process.
We discuss the model variations for the three tasks.

\paragraph{}
\textbf{Yes/No}:  The CLS token embedding from the final transformer layer is fed into a fully connected layer to obtain the logit $s$ for the binary classification. The probability of a sequence being ``yes" is given by: 
\begin{align*}
    p = \frac{1}{1+ \exp^{-(c\cdot s)}}
\end{align*}
where $c$ is the representation of the [CLS] token obtained from the final layer of BioBERT and $s$ is a learnable vector embedding. The cross entropy loss between the predicted yes probability and the corresponding ground truth is used as the loss function.
\paragraph{}
\textbf{Factoid/List}: The final layer has a start and end vector denoted by $S$ and $E$ which are trainable parameters. We compute the probabilities for the $i^{th}$ token to be the start of the answer and the $j^{th}$ token to be the end of the answer as:
\begin{align*}
    p_i^{s} = \frac{\exp^{S\cdot t_i}}{\sum_i \exp^{S \cdot t_i}},\quad p_i^{e} = \frac{\exp^{E\cdot t_i}}{\sum_i \exp^{E\cdot t_i}}
\end{align*}
where $t_i$ denotes the $i^{th}$ token's representation from the final layer of BioBERT/SciBERT. The loss is defined by taking the mean of negative log likelihood of start and end probabilities.
For selecting a span, the score of candidate span from token $t_i$ to token $t_j$ is then defined as $S\cdot t_i + E \cdot t_j$ 


\section{Leveraging Unlabeled Data for QA}
\label{sec:unsup}
Obtaining training data for Question Answering (QA) is often time-consuming and expensive. Most of existing QA datasets are only available for general domains and biomedical QA datasets are often small.
We consider approaches to leverage unlabeled text to help learn better QA models.
Recently, an approach to use unlabeled text to generate (context, question, answer) tuples for training factoid models was proposed \cite{lewis-etal-2019-unsupervised}.
However, this approach relies on a translation model to generate questions and good models for biomedical domain are not readily available.
Moreover it only applied to factoid and list type questions.
We consider this approach and introduce a simpler approach which we found to work well in practice on biomedical QA tasks.


\subsection{Unsupervised QA by cloze translation \cite{lewis-etal-2019-unsupervised}}
There are two major components of this approach.
First, a named entity extractor detects named entities from a given context. One of the occurrence of the entity is selected as the required answer for the context.
Now to create the question, the sentence in which that entity occurs is taken and the entity is replaced with a MASK token.
After this step, a cloze type question can be generated for the sentence containing the MASK token. 
Then, based on the name entity that were masked and the masked statement, a rule-based approach generates a question by selecting a wh* type words (Where, When, How, What, Wh**).
For example, if the noun phrase is a number (eg :``2020"), the question type words is most likely to be ``When". 
Since the rule based approaches can be prone to errors, the authors \cite{lewis-etal-2019-unsupervised} proposed using a seq2seq model to transform the cloze style question into a natural question.

\subsection{Self-supervised de-noising}
\begin{figure}
\begin{subfigure}{\textwidth}
  \centering
  \includegraphics[width=1\linewidth]{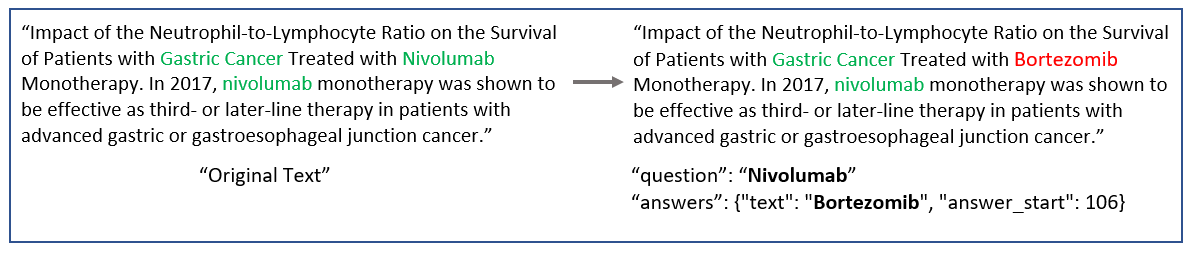}
  \caption{Factoid/List example}
  \label{fig:denoise_1}
\end{subfigure}
\begin{subfigure}{\textwidth}
  \centering
  \includegraphics[width=1\linewidth]{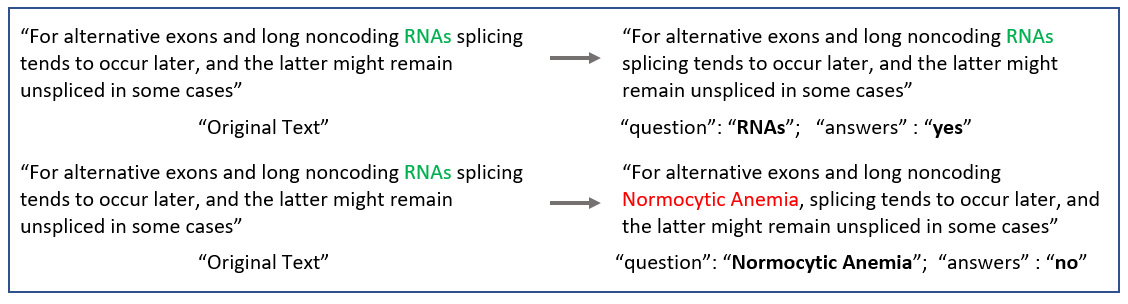}
  \caption{Yes/No example}
  \label{fig:denoise_2}
\end{subfigure}
\caption{Self-supervised de-noising approach: The original context is shown on the left, and the modified context is on the right with the entities highlighted in green and incorrect entity in red}
\label{fig:fig}
\end{figure}

The above method requires a trained seq2seq model and based on preliminary analysis we found many errors in question generation when the model was applied to biomedical text.
We thus propose here a simpler de-noising approach that doesn't require a question generation component but still helps learning better QA models from unlabeled data.
Since QA tasks often involve questions about named entities \cite{lewis-etal-2019-unsupervised}, our approach is focused on learning about mentions of entities in context by finding incorrect entity mentions in a corrupted context.
Moreover, our approach can also help learning about yes/no type of questions as we describe below.

\paragraph{Factoid/List:} The method involves corrupting a given context by randomly replacing a biomedical entity name present in the context with another entity name from same entity type. We then query the model with correct entity name as the question and the answer to this question would be the incorrect entity that it is replaced with. In the process of locating the corrupted entity location, the model would understand the semantic meaning of the context. Figure \ref{fig:denoise_1} shows an example for Factoid/List de-noising where the correct entity ``Nivolumab" is replaced by the wrong entity name ``Bortezomib" in the context (left shows the original context with all entities marked in green, and in the right we have the corrupted context with incorrect entity in red). The correct entity name is posed as the question and the model has to predict the location where its occurrence was corrupted with another entity name. 

\par
\paragraph{Yes/No:} The generation of QA data is slightly different for factoid type of questions and Yes/No type.  In case of Yes/No, we select one biomedical entity name present in the context. For a yes instance, we feed the correct biomedical entity name in the question along with the unmodified context to the model and the answer would be ``Yes". For a no instance, we replace the biomedical entity name in the context randomly with another biomedical entity, then we pair the same wrong biomedical entity name as the question and feed it to the model and the answer to this question would be a ``No". Figure \ref{fig:denoise_2} shows the example for yes/no type where on the left we have the correct context with the entity name highlighted in green and on the right we have the modified context which remains same for a ``yes" type and changed entity name in red for ``no" type.

%% file: related_work.tex
\label{sec:related}
Along with the fast development of the NLP area, QA in the biomedical domain has received much attention from the research community. 
In BioASQ 2015, Yenala et al \cite{Yenala2015IIITHAB} presented a PubMed search engine which leveraged web search results and domain words, and a new answering ranking rule to improve the question processing. 
The authors' approach relies on using the PubMed search engine to retrieve relevant documents, and then extract the snippet based on number of common domain words of the top 10 sentences of the retrieved documents and the question. 
At the same time, Zhang et al. \cite{Zhang2015AGR}
presented a generic retrieval model based on sequential dependence model, word embedding and ranking model for document retrieval. The proposed approach has two steps- split the top-ranked documents into sentences, and apply the same approach as Yenala et al \cite{Yenala2015IIITHAB} for snippets retrieval.

Lee et al. \cite{lee-etal-2016-ksanswer}
have introduced KSAnswer biomedical QA system in BioASQ 2016. 
KSAnswer was tested in the BioASQ task 4b phase A challenge. The model aims to retrieve candidate snippets using a cluster-based language model. Further, it re-ranks the retrieved top-N snippets using five independent similarity models depending on shallow semantic analysis.
SentiWordNet based lexical resource to generate the exact answers for yes/no questions was proposed \cite{inproceedings} in 2017 BioASQ challenge. The authors proposed a UMLS meta-thesaurus and term frequency metrics for answering factoid and list questions whereas a retrieval model based on UMLS concepts was used for generating ideal answers. 

Since biomedical QA datasets are usually small, many approaches have focused on generating QA data from other tasks.
EmrQA \cite{Pampari2018emrQAAL} proposed using annotated data from other clinical task by converting them into a QA format using question generation templates.
Jin et al. \cite{Jin2019PubMedQAAD} introduced a novel biomedical question answering dataset created from PubMed abstracts which involves answering a question by yes/no/maybe. This dataset contains some expert annotated data, unlabeled and artificially generated data which is used to finetune BioBERT model. The authors mention that each instance is composed of question (derived from title), context (derived from abstract), a long answer (conclusion of abstract) and yes/no/maybe summarizing the conclusion.
Since general domain QA data is abundant, transfer learning from general domain QA data, such as SQuAD \cite{Rajpurkar_2016}, has also been found beneficial for biomedical QA \cite{wiese2017neural,lee2019biobert}.

Pre-trained language models \cite{devlin2019bert,radford2019language,yang2019xlnet} have shown success in learning general purpose representations which improve performance on a number of tasks including QA \cite{devlin2019bert,lee2019biobert}.
Unsupervised approach to QA by using cloze statements as questions was proposed \cite{lewis-etal-2019-unsupervised} for general domain QA.
SpanBERT \cite{joshi2019spanbert} changed the BERT training objective to a span-based objective demonstrating improvements.
BioBERT \cite{lee2019biobert} and SciBERT \cite{beltagy2019scibert} were introduced as BERT \cite{devlin2019bert} models trained on biomedical and scientific text, respectively.
Biomedical QA using BioBERT was used \cite{yoon2019pretrained} in 2019 BioASQ Task B challenge. 
Their model outperformed previous state-of-the-art models.

Our pre-training approach requires a model for finding mentions of named entities in biomedical text. There are now increasingly accurate models that can be leveraged for this purpose \cite{wei2013pubtator,greenberg2018marginal,lee2019biobert,bansal2020simultaneously}. 
Recently, some pre-training approaches have been proposed to incorporate factual knowledge into pre-trained models \cite{peters2019knowledge,zhang2019ernie} which also require extracting such named entities from text. While these have not been applied to biomedical domain and are not focused on QA tasks, they show promise in incorporating factual knowledge in pre-trained models. Our QA targeted self-supervised de-noising method has a similar motivation and helps incorporate knowledge about biomedical named entities during pre-training.

%% file: experiments.tex
\label{sec:experiments}

\subsection{Datasets}
  We primarily use the BioASQ Task 8b-Phase B BioQA dataset to train our model and participate in the challenge. To compare our results with previous models, we train and evaluate our model on BioASQ Task 7b- Phase B BioQA dataset. The dataset contains four main types of questions - Yes/No, Factoid, List and Summary questions. The task 8b consists of 3243 question, answer pairs in the training set. The task 8b test dataset is released incrementally in 5 phases over the period of March - May 2020. Each phase has 100 test questions varying across different question types. Since, we cannot have access to the test sets to evaluate how each model we build performs, we train and evaluate our model on BioASQ Task 7b training and golden enriched test data respectively.
  The Table \ref{tab:ques_stats} shows the statistics of the question types in 7b/8b training and 7b test sets. We create two variations of the dataset. In one variation we use abstract from the documents as the context and in another version we use the snippets provided as the context. 
  
  \par
  We use additional training data to pre-train our model. We adopt techniques to generate more training data from PubMed abstracts as discussed in the earlier section of self-supervised de-noising. In case of Yes/No, more adversarial examples were created by pairing random context to questions and answering them as no type question to address class imbalance. The PubMed database has over 30 million citations for biomedical literature from MEDLINE, life science journals, and online books.
  
  \par
  Further, we use other extractive question answering dataset like SQuAD v1.1 for factoid/list type questions and SQuAD 2.0 and PubmedQA for yes/no type to train the respective model. Stanford Question Answering Dataset (SQuAD) \cite{Rajpurkar_2016} is a reading comprehension dataset consisting of over 100,000 questions posed by crowdworkers on a set of Wikipedia articles, where the answer to each question is a segment of text from the corresponding reading passage which is equivalent to a factoid answer to a question and the passage is the context. PubmedQA \cite{Jin2019PubMedQAAD} dataset with 1k expert-annotated QA consists of yes/no/maybe answers to each question. All the 'maybe' type questions are removed to match the yes/no BioASQ format. 
  We convert all datasets to one unified format which is the SQuAD dataset format where each instance has a question, context and exact answer depending on the type of question.

\begin{table}[h]
\centering
\resizebox{0.6\linewidth}{!}{
\begin{tabular}{ |c|c|c|c|c|c| } 
\hline
 Data & Total & Yes/No & List & Factoid & Summary\\ 
 \hline
 8b train & 3243 & 881 & 664 & 941 & 777\\ 
 \hline
 7b train & 2747 & 745 & 556 & 779 & 667\\
 \hline
 7b test & 500 & 140 & 88 & 162 & 110 \\
 \hline
\end{tabular}}

\caption{\label{tab:ques_stats}Dataset statistics of 8b and 7b data sets}
\end{table}

\subsection{Implementation details}
 SciBERT and BioBERT based on BERT-Base-Uncased model with 12-layer, 768-hidden, 16-heads, 340M parameter is used. Each of the datasets are pre-trained for varying no of epochs. For all models, SQuAD dataset is pre-trained for 2 epochs. A batch size of 16 or 32 with maximum sequence length of 384 is used for all the models. In case of Yes/No, model is pretrained with PubMedQA labelled dataset for 8 epochs and de-noising data for 2 epochs. The last step of training is carried out with BioASQ dataset for 2 to 4 epochs depending on other datasets the model is pre-trained on. The learning rate used for yes/no models is 5e-6 and dropout of 0.1. In case of Factoid, the learning rate used is 5e-6. The model is pretrained on de-noising data for 3 epochs and fine-tuned on BioASQ training data for 8 epochs. For List, similar configurations as factoid is exercised and set a threshold of 0.42.  
 
 \subsubsection{Evaluation Metrics}

In case of factoid, we take the highest probability answer as the exact answer. We have three evaluation metrics for factoid- Strict Accuracy, Lenient Accuracy and MRR. Strict Accuracy evaluates the exact match of the predicted answer. If the exact answer is present in the top 5 predictions then we increment the lenient score.
\input{bert_comparisons}
\input{7bresults}
\input{result.tex}

\par
In case of list, we set a threshold of 0.42 and consider all the predictions above this threshold as the list of answers to the question. The evaluation metrics for list are Precision, Recall and F1 score.
\par
In case of yesno, we use the first [CLS] from output layer and use a fully connected layer along with dropout to obtain the logit values. The positive logit values denote an `yes' while negative denotes a `no'. For each question, all the logit values for all question-context are added together and if the final value obtained is positive then its classified as an `yes' otherwise as `no'. The evaluation metrics used for the yes/no task are Accuracy, F1 score, F1 yes and F1 no scores.

\subsection{Models}

We develop different models for each type of questions. For all the tasks, our base model is either BioBERT or SciBERT. We fine tune the base model using one or more datasets chosen from SQuAD, PubMedQA, De-noising/Unsupervised Cloze Translation(UCT), PubMED, BioASQ depending on the task. The order in which the datasets are mentioned in the results Table \ref{tab:7b_results_factoid_list} is the order of fine-tuning our base model.

\par
We also use BioSentVec \cite{Chen2018BioSentVecCS}, \cite{article} model to get the embeddings and compute the similarity score between the question/context and predicted answers. We add this score to the BioBERT predicted scores to make the final predictions. In case of yes/no, a three layer neural network is trained using the embeddings to get the classification.

\subsection{Results}

In this section, we present the results of our models that are evaluated on the previous edition of BioASQ 7b, Phase B Biomedical Semantic QA challenge test dataset and on the recent BioASQ 8b, Phase B challenge. The results obtained by the different BioBERT variations on BioASQ 7b test data can be seen in Tables \ref{tab:7b_results_yes_no} and \ref{tab:7b_results_factoid_list}. A comparison of BioBERT and SciBERT are made in Tables \ref{tab:bio_sci_yesno} and \ref{tab:7b_results_factoid_list}. The results obtained from the leader board of BioASQ 8b challenge \cite{BioASQ8b} is shown in Table \ref{tab:8b_results}.

\subsubsection{Results on BioASQ 7b test sets}
Various experiments are performed and performance of the models are evaluated for the task of question answering that includes yes/no, factoid and list type of questions. The 7b test batches are used for evaluating the model performance.  The results are shown in Tables \ref{tab:7b_results_yes_no} and \ref{tab:7b_results_factoid_list}. We reproduced the previous winners results using our base model, BioBERT with SQuAD dataset \cite{yoon2019pretrained}

 The general trend observed is that adding more data tends to improve the performance of the model. This can be clearly seen from data illustrated in the Tables \ref{tab:7b_results_yes_no} and \ref{tab:7b_results_factoid_list}. Even when fine-tuned with additional non-biomedical dataset such as SQuAD, the model is able to perform better than baseline. The use of PubmedQA  along with SQuAD datasets for fine-tuning made the Yes/No model more robust.

  It is observed that training the model with unsupervised and self supervised data before fine-tuning with BioASQ data presents a significant boost to the test performance. Unsupervised data generation approach led to increase in model performance for list and factoid. The performance boost can be seen in-spite of the fact that the data generated by unsupervised approach was not of particularly high quality. It can be attributed to the improvement of the model's generalization ability with the huge amount of data. We can also note that using the BioSentVec in the list and factoid along with BioBERT also aids the results. For the factoid and list type question,  the BioSentVec is used to generate the similarity scores and use it to balance the power of BioBERT logit in the last layer. The gain obtained on the testing set is consistent with our hypothesis that the correct answer should be similar to the question semantically.  
 
  It can be seen that when the model is trained with self supervised de-noising approach it either out performed all the variations tried by significant margin or matched unsupervised approach for 7b test. In all three question types, the increase in performance can be clearly seen. In case of Yes/No type, the performance of the model greatly improved by the de-noising the data. However, it is noticed that having too many datasets for fine-tuning with de-noising approach resulted in slight decrease in the performance comparatively. The key takeaway is that self supervised approach performed well with less fine-tuning data and trained faster with fewer epochs than other approaches.  
 
 A comparison is carried out to analyse performance of BioBERT and SciBERT models in comparable data settings and no significant difference in performance is observed. Tables \ref{tab:7b_results_factoid_list}  and \ref{tab:7b_results_yes_no} clearly show that there is no much difference in performance between SciBERT and BioBERT. In case of Yes/No with only BioASQ training, it can be seen that the accuracy of SciBERT is higher. This is mainly due to the large imbalance in the test set towards the positive class ("yes") and the F1 scores are almost the same for both BioBERT and SciBERT with BioASQ training alone. 

\subsubsection{Results on BioASQ 8b Challenge}
The Table \ref{tab:8b_results} reports the results obtained on the BioASQ 8b challenge which is conducted in 5 phases.  The results are put up on the leaderboard \cite{BioASQ8b}. We show the comparison of our model performance with the best performing team on the leaderboard for each batch and in each of three type of questions. The challenge ranks the performance based on the Yes/No Accuracy, Factoid- MRR and List- F Measure metrics. The observations made from the Tables \ref{tab:7b_results_yes_no} and \ref{tab:7b_results_factoid_list} can be validated from the 8b challenge results.

 In each of the different phases, models with different configurations are submitted for evaluation. 
 In case of Yes/No Question type, for the test batch 1 BioBERT model fine-tuned on BioASQ 8b training dataset is used. In case of test batch 2, the submission is made by adding in SQuAD fine-tuning to the BioBERT model. For the 3rd and 4th test phases, the submission is made with fine-tuning on PubmedQA and SQuAD datasets. The $5^{th}$ test phase submission is carried out with our best model which is with de-noising method.
 
\par 
In case of Factoid, for the test batch 1, the BioBERT model fine-tuned on SQuAD and BioASQ 8b training data is submitted. For test batch 2, we modified the 8b training data by adding a start index of the exact answer while fine-tuning the model. The BioBERT model fine-tuned with unsupervised data is submitted in the test batch 3 while SciBERT base model and with SQuAD is submitted in test batch 4. The last and final submission is made with BioBERT model pretrained on de-noising data.
\par
In case of List, the model variations follow as mentioned in Factoid. We submitted BioBERT fine-tuned on SQuAD and BioASQ 8b data for first two test batches with modifications in computing the exact answer start for the second batch. The BioBERT model fine-tuned with unsupervised data is submitted in the test batch 3. BioBERT and SciBERT models fine-tuned on SQuAD and unsupervised data are in submitted test batch 4. The $5^{th}$ test phase submission is carried out with our best model that includes training with de-noising data. 

Interestingly, unlike the trend observed in 7b test results, Table \ref{tab:7b_results_factoid_list} for self-supervised de-noising and unsupervised cloze-translation approaches, pre-training with de-noising approach gave better MRR and F-Measure for factoid and list respectively as shown in Table \ref{tab:8b_comp}. The BioASQ 8b test batch 5 gave an MRR=0.6354 for de-noising approach while it is only 0.5604 for UCT approach in case of factoid. Similarly for list, we achieved F-measure=0.3353 using de-noising data approach while it is 0.2166 for unsupervised approach. 

%% file: bert_comparisons.tex
\begin{table*}[t]
    \centering
    \resizebox{0.8\textwidth}{!}{
    \begin{tabular}{|c|c|c|c|c|c|}
     \hline
Dataset & \multicolumn{5}{|c|}{Yes/No} \\
\cline{2-6}  & Variant & Acc &   F1 Score & F1 Yes &F1 No   \\
\hline
BioASQ & BioBERT &  0.76 & 0.76 & 0.78 & 0.74\\
\cline{2-6} & SciBERT  &  0.80 & 0.77 & 0.85 & 0.70 \\
\hline
PubMedQA + BioASQ & BioBERT  & 0.83 & 0.81 & 0.87 & 0.76  \\
\cline{2-6} & SciBERT & 0.83& 0.82& 0.86& 0.78 \\
\hline
\end{tabular}}
\captionsetup{width=10cm}
\caption{ Performance  comparison  between BioBERT vs SciBERT  for Yes/No question answers using different datasets and tested on BioASQ 7b test data.}
\label{tab:bio_sci_yesno}
\end{table*}

\begin{table*}[t]
    \centering
    \resizebox{0.8\textwidth}{!}{
    \begin{tabular}{|c|c|c|c|c|c|c|}
     \hline
Dataset  & \multicolumn{3}{|c|}{Factoid} & \multicolumn{3}{|c|}{List} \\
\cline{2-7}  & Variant & SAcc & LAcc   & Precision & Recall &MacroF1 \\
\hline
BioASQ & BioBERT &  0.23 & 0.38 & 0.55 & 0.32 & 0.38 \\
\cline{2-7}  & SciBERT & 0.23&0.36&0.49 &0.28 &0.33 \\
\hline
SQuAD + BioASQ &  BioBERT &   0.26 & 0.49 &  0.60 & 0.45 &0.48\\
\cline{2-7} & SciBERT & 0.26 & 0.49 & 0.61 & 0.47 & 0.49\\
\hline
\end{tabular}}
\captionsetup{width=10cm}
\caption{Performance comparison between BioBERT vs SciBERT  for Factoid/List question answers using different datasets and tested on BioASQ 7b test data.}
\label{tab:bio_sci_factoidlist}
\end{table*}

%% file: 7bresults.tex
\begin{table*}[t]
    \centering
    \resizebox{\textwidth}{!}{
    
\begin{tabular}{|l|c|c|c|c|}
\hline
\hspace{1.5in}Model & \multicolumn{4}{|c|}{Yes/No} \\ 
\hline
 & Acc & F1Score & F1 Yes &F1 No  \\
\hline 
BioBERT - BioASQ &  0.76 & 0.76 & 0.78 & 0.74\\
\hline
BioBERT- SQuAD + BioASQ &  0.80 & 0.78& 0.84&0.7 \\
\hline
BioBERT- PubMedQA + BioASQ & 0.83 & 0.81 & 0.87 & 0.76 \\
\hline
BioBERT- SQuAD + Denoising + BioASQ & \textbf{0.88} & \textbf{0.88} & \textbf{0.90} & \textbf{0.86}\\
\hline
BioBERT- SQuAD + PubMedQA + BioASQ & 0.83& 0.81 & 0.87 & 0.76  \\
\hline
BioBERT- SQuAD + PubMedQA + Denoising + BioASQ & 0.86 & 0.86 & 0.88 &0.73\\
\hline
BioSentVec - 3layer NN + BioASQ & 0.66 & 0.64 & 0.74 & 0.55 \\
\hline
\end{tabular}}
\caption{Performance comparison between different variations of models for Yes/No question answers on BioASQ 7b test data }
\label{tab:7b_results_yes_no}
\end{table*}

\begin{table*}[t]
    \centering
    \resizebox{\textwidth}{!}{
    \begin{tabular}{|l|c|c|c|c|c|}
     \hline
\hspace{1.5in}Model & \multicolumn{2}{|c|}{Factoid} & \multicolumn{3}{|c|}{List} \\
\hline
& SAcc & LAcc & Precision & Recall &MacroF1 \\
\hline
BioBERT - BioASQ &  0.23 & 0.38 & 0.55 & 0.32 & 0.38\\
\hline
BioBERT- SQuAD + BioASQ &   0.26 & 0.49 &  0.60 & 0.45 &0.48\\
\hline
BioBERT- SQuAD + Denoising + BioASQ &\textbf{0.28}&\textbf{0.54}& 0.68 & \textbf{0.49}& \textbf{0.51}\\
\hline
BioBERT- SQuAD + UCT + BioASQ  & 0.28 & 0.54 & \textbf{0.70} & 0.46 & 0.50 \\
\hline
BioBERT + BioSentVec- SQuAD + UCT  + BioASQ & 0.31 & 0.58 & 0.73 & 0.35 & 0.51 \\
\hline

\end{tabular}}
\caption{Performance comparison between different variations of models for Factoid and List type of question answers on BioASQ 7b test data }
\label{tab:7b_results_factoid_list}
\end{table*}

%% file: result.tex
\begin{table*}[t]
    \centering
    \resizebox{\textwidth}{!}{
    \begin{tabular}{|c|c|c|c|c|c|c|c|c|c|c|c|c|c|}
     \hline
Batch & \multicolumn{5}{|c|}{Yes/No} & \multicolumn{4}{|c|}{Factoid} & \multicolumn{4}{|c|}{List} \\
\cline{2-14}  & System & Acc &   F1 Yes &F1 No & F1 Score & System & SAcc & LAcc & MRR & System & Precision & Recall &Macro F1 \\
\hline
1 & Ours & 0.6800 & 0.7778 & 0.4286 & 0.6032 & Ours & 0.3750 & 0.5938 & \textbf{0.4688} & Ours & 0.4875 & 0.2983 & 0.3448 \\
\cline{2-14} & Others & \textbf{0.8800} & 0.9091 & 0.8235 &0.8663 & Others & 0.3438 & 0.6250 & 0.4583 & Others & 0.3884 & 0.5629 & \textbf{0.4315}\\
\hline
2 & Ours & 0.7778 & 0.8519 & 0.5556 & 0.7037 & Ours & 0.1200 & 0.2000 & 0.1480 & Ours & - & - & - \\
\cline{2-14} & Others & \textbf{0.9444} & 0.9630 & 0.8889 &0.9259 & Others & 0.2800 & 0.4400 & \textbf{0.3533} & Others & 0.5643 & 0.4643 & \textbf{0.4735}\\
\hline
3 & Ours & \textbf{0.9032} & 0.9143 & 0.8889 & 0.9016 & Ours & 0.3214 & 0.4643 & 0.3810 & Ours & 0.7361 & 0.4833 & \textbf{0.5229} \\
\cline{2-14} & Others & 0.9032 & 0.9189 & 0.8800 &0.8995 & Others & 0.3214 & 0.5357 & \textbf{0.3970} & Others & 0.5278 & 0.4778 & 0.4585\\
\hline
4 & Ours & 0.8077 & 0.8387 & 0.7619 & 0.8003 & Ours & 0.5000 & 0.7059 & 0.5637 & Ours & 0.5753 & 0.4182 & 0.4146 \\
\cline{2-14} & Others & \textbf{0.8462} & 0.8571 & 0.8333 &0.8452 & Others & 0.5588 & 0.7353 & \textbf{0.6384} & Others & 0.5375 & 0.5089 & \textbf{0.4571}\\
\hline
5 & Ours & 0.7941 & 0.8205 & 0.7586 & 0.7896 & Ours & 0.5625 & 0.7188 & \textbf{0.6354} & Ours & 0.5972 & 0.3819 & 0.4421 \\
\cline{2-14} & Others & \textbf{0.8529}	& 0.8649 & 0.8387 & 0.8518 & Others & 0.5313 & 0.7188 & 0.6120 & Others & 0.5516 & 0.5972 & \textbf{0.5618}\\
\hline
\end{tabular}}
\caption{BioASQ challenge 8b results per batch. The table here shows our results and the best results among other teams in the leader board. }
\label{tab:8b_results}
\end{table*}

\begin{table}[h!]
    \centering
    \resizebox{\textwidth}{!}{
    \begin{tabular}{|c|c|c|c|c|c|c|c|}
    \hline
    Model & Dataset & \multicolumn{3}{|c|}{Factoid} & \multicolumn{3}{|c|}{List}\\
    \cline{3-8} & & SAcc & LAcc & MRR & Precision & Recall & Macro F1\\
    \hline
    BioBERT & SQuAD + UCT + BioASQ & 0.4688 & \textbf{0.7188} & 0.5604 & 0.3750 & 0.1756 & 0.2166\\
    \hline
    BioBERT & SQuAD + Denoising + BioASQ & \textbf{0.5625} & \textbf{0.7188} & \textbf{0.6354} & \textbf{0.5139} & \textbf{0.2808} & \textbf{0.3353}\\
    \hline
    \end{tabular}}
    \caption{Results of our different models on BioASQ Task 8b Phase B test batch 5 dataset}
    \label{tab:8b_comp}
\end{table}

%% file: conclusion.tex
\label{sec:conclusion}
In this work we evaluated pre-trained models, BioBERT and SciBERT, for biomedical QA.
We proposed a novel approach to pre-training, self-supervised de-noising, which enables learning good representations for QA tasks.
The experimental results show that the approach improves the performance of the models in all the QA tasks. One main advantage of this approach is that it is simple and does not require expensive annotation, enabling large-scale pre-training.
In the future, it will be interesting to extend the approach to include more complex reasoning required for QA, for example reasoning about multiple entities in context, and including natural form questions generated either through templates or a learned model.

%% file: acknowledgement.tex
This work was supported in part by the UMass Amherst Center for Data Science and the Center for Intelligent Information Retrieval, in part by the Chan Zuckerberg Initiative, and in part by the National Science Foundation under Grant No. IIS-1514053 and IIS-1763618. Any opinions, findings and conclusions or recommendations expressed in this material are those of the authors and do not necessarily reflect those of the sponsor.